\documentclass[10pt, a4paper]{article}

\usepackage[final]{lrec-coling2024} 

\usepackage{multirow}
\usepackage{amsmath}
\usepackage{amssymb}
\usepackage[subrefformat=parens]{subcaption}
\usepackage{algorithm}
\usepackage{algpseudocode}
\usepackage{booktabs}
\usepackage{float}
\usepackage{pifont}

\newcommand{\cmark}{\ding{51}}%
\newcommand{\xmark}{\ding{55}}%

\newcommand{\mr}[2]{\multirow{#1}{*}{#2}}
\newcommand{\ml}[3]{\multicolumn{#1}{#2}{#3}}

\title{JMultiWOZ: A Large-Scale Japanese Multi-Domain Task-Oriented Dialogue Dataset}

\name{Atsumoto Ohashi\textsuperscript{*}\thanks{* Equal contribution}, Ryu Hirai\textsuperscript{*}, Shinya Iizuka, Ryuichiro Higashinaka} 

\address{Graduate School of Informatics, Nagoya University, Japan \\
         \{ohashi.atsumoto.c0, hirai.ryu.k6, iizuka.shinya.a8\}@s.mail.nagoya-u.ac.jp\\
         higashinaka@i.nagoya-u.ac.jp}

\abstract{
Dialogue datasets are crucial for deep learning-based task-oriented dialogue system research. While numerous English language multi-domain task-oriented dialogue datasets have been developed and contributed to significant advancements in task-oriented dialogue systems, such a dataset does not exist in Japanese, and research in this area is limited compared to that in English. In this study, towards the advancement of research and development of task-oriented dialogue systems in Japanese, we constructed JMultiWOZ, the first Japanese language large-scale multi-domain task-oriented dialogue dataset. Using JMultiWOZ, we evaluated the dialogue state tracking and response generation capabilities of the state-of-the-art methods on the existing major English benchmark dataset MultiWOZ2.2 and the latest large language model (LLM)-based methods. Our evaluation results demonstrated that JMultiWOZ provides a benchmark that is on par with MultiWOZ2.2. In addition, through evaluation experiments of interactive dialogues with the models and human participants, we identified limitations in the task completion capabilities of LLMs in Japanese.
\\ \newline \Keywords{Multi-domain task-oriented dialogue, Dialogue state tracking, Response generation} }

\begin{document}

\maketitleabstract

\section{Introduction}
Methods based on deep learning have been actively introduced in the research of task-oriented dialogue systems~\citep{gao2018neural, zhang2020recent}, which have greatly improved their performance on task completion~\citep{Zhang_Ou_Yu_2020, NEURIPS2020_e9462095, He_Dai_Zheng_Wu_Cao_Liu_Jiang_Yang_Huang_Si_Sun_Li_2022}. Task-oriented dialogue datasets are essential for developing these neural models, and a number of single-domain task-oriented dialogue datasets in English have been developed previously~\citep{henderson-etal-2014-second, wen-etal-2017-network, eric-etal-2017-key, shah2018building}.

MultiWOZ is a large-scale dialogue corpus that was developed to address more complex multi-domain dialogues~\citep{budzianowski-etal-2018-multiwoz}. It includes dialogues spanning seven domains, namely, tourist attractions, hotels, restaurants, taxis, trains, police stations, and hospitals, and led to the development of subsequent dialogue models. Following the introduction of MultiWOZ, numerous large-scale dialogue datasets have been constructed~\citep{Rastogi_Zang_Sunkara_Gupta_Khaitan_2020, mosig2020star, chen-etal-2021-action}, and task-oriented dialogue models using these as benchmarks have been actively researched. CrossWOZ~\citep{zhu-etal-2020-crosswoz} and other large-scale multi-domain dialogue datasets have been introduced for Chinese as well~\citep{quan-etal-2020-risawoz, dai-etal-2022-cgodial}, promoting research on Chinese task-oriented dialogue systems.

\begin{figure}
\centering
\includegraphics[width=1\linewidth]{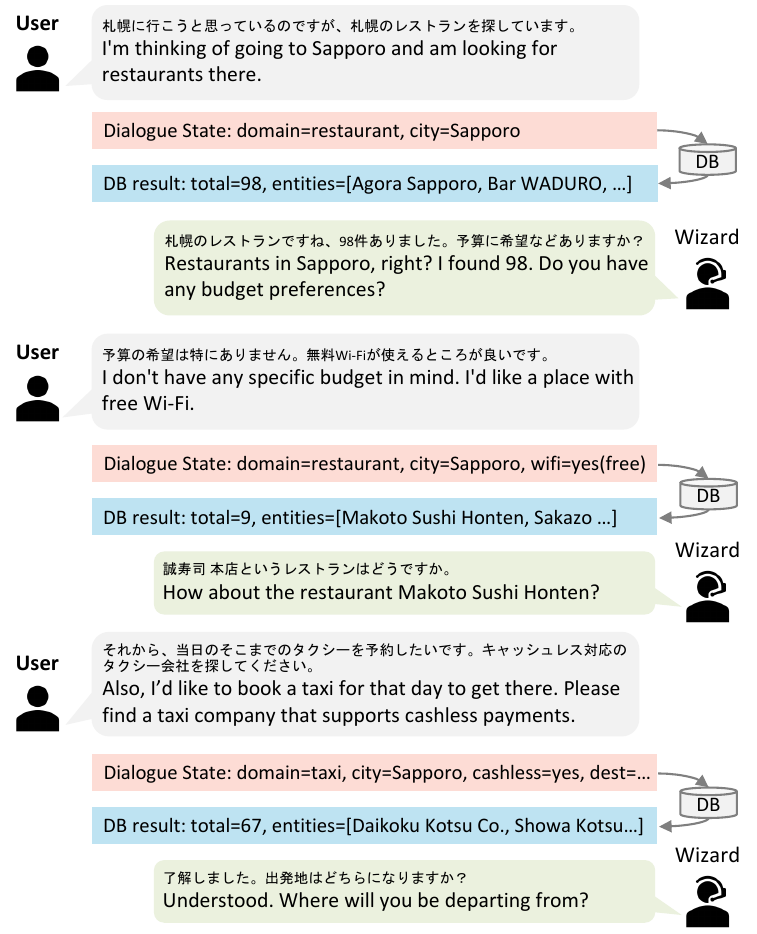}
\caption{An example of dialogue across two domains: restaurants and taxis. The gray and green message bubbles represent the utterances of the user and the wizard, respectively. The red and blue boxes indicate the annotation of the dialogue state and the database results, respectively. The bubble for each utterance contains both the original Japanese utterance and its English translation by the authors.}
\label{fig:dialogue_example}
\end{figure}

However, due to the high cost of constructing task-oriented dialogue corpora, there are fewer multi-domain task-oriented dialogue corpora in other languages compared to English and Chinese~\citep{hung-etal-2022-multi2woz}. In this study, we focus on Japanese, in which the research and development of task-oriented dialogue models based on deep learning have been limited thus far.

Toward the research and development of task-oriented dialogue systems in Japanese, we have constructed the Japanese Multi-Domain Wizard of Oz (JMultiWOZ), the first Japanese multi-domain task-oriented dialogue dataset.\footnote{Our data and code are publicly available at \url{https://github.com/nu-dialogue/jmultiwoz}} JMultiWOZ contains a total of 4,246 conversations spanning six travel-related domains (tourist attractions, accommodation, restaurants, shopping facilities, taxis, and weather). An example of a dialogue is shown in Figure~\ref{fig:dialogue_example}. JMultiWOZ provides the dialogue state at each turn and the database of each domain for implementing and benchmarking task-oriented dialogue models.

In this paper, we outline the procedure for constructing the JMultiWOZ dataset and introduce its statistics. We evaluated the dataset on the two main tasks of task-oriented dialogue enabled by JMultiWOZ, i.e., dialogue state tracking (DST) and response generation (RG), using the state-of-the-art (SOTA) methods~\citep{bang-etal-2023-task} and the latest LLM-based methods~\citep{hudecek-dusek-2023-large}. For further validation, the end-to-end dialogue capability of these dialogue models was evaluated by interactions with human participants. The contributions of this study are threefold:
\begin{itemize}
\item We constructed JMultiWOZ, the first large-scale Japanese multi-domain task-oriented dialogue dataset.
\item We evaluated the dataset on DST and RG tasks using existing SOTA models and the latest LLMs, and demonstrated that JMultiWOZ can provide a Japanese benchmark of complexity comparable to that of the major English dataset MultiWOZ2.2.
\item We conducted a human evaluation experiment, which showed that, even with the latest LLMs, there remain challenges concerning the capabilities of task-oriented dialogue in Japanese.
\end{itemize}

\begin{table*}[t]
\footnotesize
\centering
\begin{tabular}{l|p{12.5cm}} \toprule
\textbf{Domain} & \textbf{Slots} \\ \midrule
restaurant & city, name, genre, area, pricerange, station, wifi, parking, opentime, phone, address, accesstime, closed, priceinfo, reference, people, day, time \\ \midrule
accommodation & city, name, genre, area, pricerange, station, wifi, parking, withrestaurant, phone, address, accesstime, checkintime, checkouttime, priceinfo, reference, people, day, stay \\ \midrule
attraction & city, name, genre, area, station, wifi, parking, opentime, phone, address, accesstime, closed, adultfee, childfee, priceinfo \\ \midrule
shopping & city, name, genre, area, station, parking, opentime, phone, address, accesstime, closed \\ \midrule
taxi & city, name, cashless, jumbo, phone, reference, day, time, departurepoint, arrivalpoint \\ \midrule
weather & city, area, day, weather, mintemperature, maxtemperature \\ \bottomrule
\end{tabular}
\caption{List of slots defined in the ontology for each domain}
\label{tbl:ontology}
\end{table*}

\section{Related Work}
\subsection{Task-Oriented Dialogue Corpora}
Many task-oriented dialogue corpora have been created in English thus far. Previously, single-domain dialogues, where only one domain appears in a dialogue, were predominant, such as WOZ2.0~\citep{wen-etal-2017-network}, Frames~\citep{el-asri-etal-2017-frames}, and KVRET~\citep{eric-etal-2017-key}. All of these dialogues were conducted using the Wizard of Oz (WOZ) method~\citep{Kelley1984AnID}, with a human user and another person acting as the system (i.e., the wizard). There are also human-to-machine~\citep{henderson-etal-2014-second} and machine-to-machine dialogue corpora simulated between machines~\citep{shah2018building}. Multi-domain dialogue corpora, in which multiple domains appear in a single dialogue, are being increasingly constructed to address more complex requirements. MultiWOZ~\citep{budzianowski-etal-2018-multiwoz} is a representative example, being a large-scale corpus with over 10,000 dialogues covering seven travel-related domains. Other existing large-scale multi-domain dialogue corpora include Schema-Guided Dialogue~\citep{Rastogi_Zang_Sunkara_Gupta_Khaitan_2020}, STAR~\citep{mosig2020star}, and ABCD~\citep{chen-etal-2021-action}.

There are also several large-scale multi-domain dialogue corpora in Chinese. CrossWOZ~\citep{zhu-etal-2020-crosswoz} is the first multi-domain dialogue corpus in Chinese and contains about 6,000 travel-related dialogues. The RiSAWOZ corpus~\citep{quan-etal-2020-risawoz} introduced more domains and dialogues, and subsequently, CGoDial~\citep{dai-etal-2022-cgodial} was devised as an extension of other dialogue corpora including RiSAWOZ. In addition, BiTOD~\citep{lin2021bitod} was constructed to develop bilingual multi-domain dialogue models in both English and Chinese.

Constructing task-oriented dialogue corpora is generally costly, and there are few large-scale multi-domain datasets outside of English and Chinese~\citep{hung-etal-2022-multi2woz}. SCUD~\citep{hayashibe2022self} is a Japanese single-domain task-oriented dialogue corpus related to accommodation search but it only contains 210 dialogues.

\subsection{Translation-based Corpora}
Given the high cost of constructing task-oriented dialogue corpora, efforts are being made to construct dialogue corpora in other languages by translating readily available English corpora. For instance, GlobalWOZ~\citep{ding-etal-2022-globalwoz} is a dialogue corpus that expanded MultiWOZ into 17 languages through machine translation. Among the 17 languages, high quality has been achieved in three languages (Chinese, Spanish, and Indonesian) through post-editing by professional translators for some dialogues in the test set. However, the quality is not guaranteed for other languages, including Japanese. Other corpora constructed from machine translation and manual post-editing of MultiWOZ include AllWOZ~\citep{zuo2021allwoz} and Multi$^2$WOZ~\citep{hung-etal-2022-multi2woz}, neither of which includes Japanese.

Problems due to poor translations have been reported in translation-based dialogue corpora (e.g., `translationese,' lack of cultural adaptation)~\citep{majewska-etal-2023-cross}, which may prevent the models' practical performance from being evaluated accurately~\citep{hu-etal-2023-multi-3}. Therefore, this study aims to construct a realistic dataset in the Japanese context by collecting dialogues from scratch.

\section{Data Collection}
JMultiWOZ is a corpus containing dialogues of travelers planning a trip to one of nine cities in Japan (Sapporo, Sendai, Tokyo, Yokohama, Nagoya, Kyoto, Osaka, Fukuoka, and Naha) while collecting tourist information. The six domains include tourist attractions, accommodations, restaurants, shopping facilities, taxis, and weather. Using the WOZ method, each dialogue was conducted by two human interlocutors, one as a traveler (user) and the other as an information provider (wizard).

The following sections describe the five steps of constructing this corpus: (1) ontology definition, (2) construction of the backend database that the wizard uses to obtain travel information, (3) design and creation of user goals, (4) dialogue collection, and (5) annotation of the full dialogue state.

\subsection{Definition of Ontology}
\label{sec:definition_of_ontology}
In a task-oriented dialogue, the \emph{ontology} represents the structure of the backend database. Specifically, it defines attributes such as name, address, and phone number for each \emph{entity} in the database. An entity is a unit of record in the database, such as a specific tourist attraction or restaurant, and its attributes are called \emph{slots}. In this study, the ontology of each domain was defined with reference to existing studies~\cite{budzianowski-etal-2018-multiwoz, zhu-etal-2020-crosswoz} while considering the characteristics of Japanese culture so as not to be unnatural. For instance, dialogues related to the police and hospital domains that exist in MultiWOZ are rarely encountered in Japanese dialogue travel centers. In view of this, we eliminated police and hospital domains, and instead introduced more culturally appropriate domains such as shopping and weather. The ontology for all domains is shown in Table~\ref{tbl:ontology}.

\subsection{Construction of Backend Database}
Based on the ontology, a backend database that the wizard uses to retrieve entities and travel information during the dialogue was constructed for each domain. To enhance the realism of the dialogue using real entities, lists of facilities publicly available from governments and municipalities of various cities were used to construct databases for tourist spots, accommodations, restaurants, shopping facilities, and taxis. From this list, only facilities that have publicly accessible websites were selected to be included in the database, and information for all slots of each entity was manually obtained from the website. For the taxi domain, the unit of entities was set as taxi companies.

The final number of entities contained in the database for each domain were as follows: 447 for tourist spots, 884 for accommodations, 952 for restaurants, 445 for shopping facilities, and 167 for taxis. For the construction of the weather domain database, the unit of entity was set as the date, and 365 days' worth of weather information was artificially created for each city.

\begin{table}[t]
\footnotesize
\centering
\begin{tabular}{lll} \toprule
\textbf{Domain} & \ml{2}{l}{\textbf{Slot}} \\ \midrule
\mr{4}{attraction} & \mr{3}{info.} & city = Nagoya \\
 & & parking = yes(free) \\
 & & wifi = yes(free) \\ \cmidrule{2-3}
 & reqt. & area, station, closed \\ \midrule
\mr{3}{accommodation} & \mr{2}{info.} & pricerange = expensive \\
 & & wifi = yes(free) \\ \cmidrule{2-3}
 & reqt. & accesstime, genre \\ \bottomrule
\end{tabular}
\caption{Example of a user goal. ``info.'' and ``reqt.'' in the ``Slot'' column indicate informable and requestable slots, respectively.}
\label{tbl:example_of_goal}
\end{table}

\begin{figure*}
\centering
\includegraphics[width=0.85\linewidth]{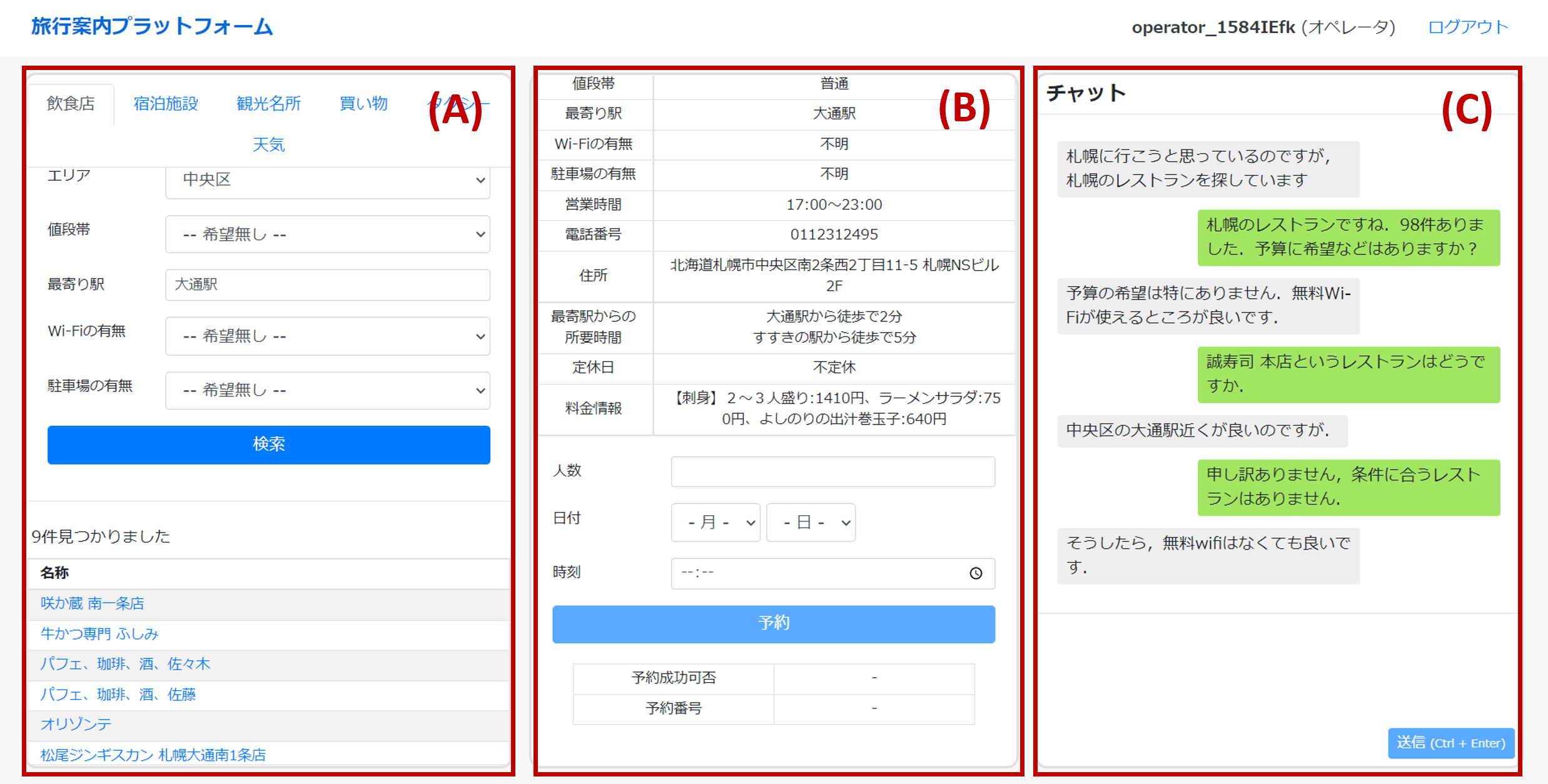}
\caption{Web UI of wizard for dialogue collection: (A): A search form for an entity from the backend database, (B): An interface displaying detailed information about the selected entity and a reservation form for the entity, (C): An interface for chatting with the user.}
\label{fig:ui_system_chat_with_frames}
\end{figure*}

\subsection{Design of User Goal}
A user goal is the objective the user aims to achieve through the dialogue with the wizard, and one goal is set for each dialogue. An example of a user goal is shown in Table~\ref{tbl:example_of_goal}. Each goal covers one or more domains, and each domain in the goal consists of one or more \emph{informable} slots, which are the search criteria for the entity the user is looking for, such as the desired budget or destination, and one or more \emph{requestable} slots, which are the attribute information of the entity the user needs to obtain, such as phone numbers or addresses. In domains where reservations are commonly made in real life (for instance, after finding the desired restaurant or accommodation, reservations are often made with conditions such as the date and number of people), several conditions for reservation are randomly added as \emph{book} slots.

The composition and diversity of user goals are directly linked to the naturalness and diversity of the dialogues in the corpus. To introduce diversity in dialogue length and complexity, 1-3 domains are randomly selected for each user goal. Next, slots to be included in each domain are randomly chosen. From the ontology defined in Section \ref{sec:definition_of_ontology}, 2-7 slots, including informable, requestable, and possibly bookable slots, are selected for each domain. The informable slot ``city'', which indicates the user's tourist destination city, is shared among the domains within a user goal. To enhance the realism of the dialogue, following \citep{budzianowski-etal-2018-multiwoz}, some goals were set with tasks to change the value of an informable slot and/or book slot amid the dialogue (for instance, change the originally communicated reservation condition of 5 p.m. to 6 p.m.). Ahead of subsequent dialogue collection, a total of 5,000 unique user goals were created.

\subsection{Dialogue Collection}
\label{sec:dialogue_collection}
Dialogues were collected using a backend database and randomly generated dialogue goals. Based on the dialogue collection platform\footnote{\url{https://github.com/thu-coai/CrossWOZ}} of \citep{zhu-etal-2020-crosswoz}, the dialogue web UIs for the user and wizard were implemented. The web UI used by the wizard is depicted in Figure~\ref{fig:ui_system_chat_with_frames}. Crowd workers for both the user and wizard roles were recruited via Lancers\footnote{\url{https://www.lancers.jp}}, a major Japanese crowdsourcing service. Only those who consented to the publication of data obtained from the dialogue collection participated.

The workers read the instructions in the dialogue manual, watched a demonstration video that explained how to operate the web UI, and learned the workflow before participating in the dialogues. To ensure diverse user utterances, each user could participate in a maximum of 100 dialogues, and they could engage with the same wizard for a maximum of 20 dialogues. Meanwhile, because it is preferable for the wizard to behave consistently, there was no limit to the number of dialogues, and the same wizard was allowed to engage in conversations repeatedly. In the end, 65 users and 18 operators participated in the dialogue collection. The tasks of the user and wizard, and the quality control for the wizards are explained in detail in the following paragraphs.

\begin{table}[t]
\small
\centering
\definecolor{MyBlue}{HTML}{0041ff}
\definecolor{MyGreen}{HTML}{35a16b}
\begin{tabular}{lp{6.5cm}}
\toprule
1. & You are planning a trip to \textcolor{MyBlue}{\textbf{Nagoya}}. \\
2. & Find a \textcolor{MyBlue}{\textbf{tourist attraction}} to visit for one day. Choose a place where the parking is \textcolor{MyBlue}{\textbf{free}}. The place should also have \textcolor{MyBlue}{\textbf{free}} Wi-Fi.\\
3. & Once you find a desired tourist spot, ask about its \textcolor{MyGreen}{\textbf{area, nearest station, and closed days}}.\\
4. & Find a \textcolor{MyBlue}{\textbf{place to stay}} for the night. The budget is on the \textcolor{MyBlue}{\textbf{higher}} side. The place should also offer \textcolor{MyBlue}{\textbf{free}} Wi-Fi.\\
5. & Once you find a suitable accommodation, ask about the \textcolor{MyGreen}{\textbf{time it takes from the nearest station and the type of accommodation}}. \\
\bottomrule
\end{tabular}
\caption{Explanation of the user goal shown in Table~\ref{tbl:example_of_goal}. The \textcolor{MyBlue}{\textbf{blue text}} indicates informable slot values, while the \textcolor{MyGreen}{\textbf{green text}} indicates requestable slot values.}
\label{tbl:example_of_goal_description}
\end{table}

\paragraph{User's Task} The user aims to properly convey the informable slots in the user goal to the wizard, obtain information about the requestable slots from the wizard, and in some cases, reserve the entity with the conditions of booking slots. Instead of a user goal formatted as in Table~\ref{tbl:example_of_goal}, users comprehend the goal by reading template sentences that describe each slot. Template sentences that explain the user goal in Table~\ref{tbl:example_of_goal} are shown in Table~\ref{tbl:example_of_goal_description}.

\paragraph{Wizard's Task} The wizard searches for entities in the backend database that match the constraints conveyed by the user and provides the user with information about the found entities. The wizard's web UI (Figure~\ref{fig:ui_system_chat_with_frames}) has interfaces for (A) searching for entities based on the user's criteria and (B) checking detailed information or making reservations for the found entities in addition to (C) the panel for chatting with the user. The database search query (DB query) input by the wizard in each turn is recorded as part of the dialogue state (details can be found in Section \ref{sec:annotation_of_full_dialogue_state}).

\paragraph{Quality Control of Wizard} Some studies~\citep{eric-etal-2020-multiwoz, zang-etal-2020-multiwoz} have reported issues such as inconsistencies in the slot value notation within the dialogue state (e.g., the value ``18:00'' for the time slot appears in several ways: ``1800'', ``0600pm'', and ``6 PM'') in existing corpora. Such inconsistencies can confuse or underestimate the capabilities of the dialogue model, making it impossible to provide an appropriate benchmark. Therefore, input values were selected from dropdown menus to prevent wizards from manually entering DB queries. Additionally, to enhance the quality of the wizard, 3-5 practice dialogues were conducted beforehand. The workers received feedback from the authors on errors. Only workers who no longer had issues after repeated feedback participated in the actual dialogues.

\begin{table}[t]
\small
\centering
{\tabcolsep=1.5mm
\begin{tabular}{lrrrr}
\toprule
& \textbf{Total} & \textbf{Train} & \textbf{Dev} & \textbf{Test} \\
\midrule
Dialogues & 4,246 & 3,654 & 300 & 300 \\
Turns & 61,186 & 52,405 & 4,346 & 4,435 \\
Tokens & 1,102,658 & 943,653 & 78,683 & 80,322 \\
Vocab. & 11,121 & 10,309 & 2,854 & 2,971 \\
\bottomrule
\end{tabular}
}
\caption{Statistics of JMultiWOZ}
\label{tbl:statistics_of_jmultiwoz}
\end{table}

\vskip\baselineskip
Through the above procedure, a total of 4,508 dialogues were collected. Dialogues that would be considered noise in the dataset were then revised or excluded through the following two procedures:
\begin{enumerate}
\item At the end of each dialogue, workers completed questionnaires to report any issues that occurred during the dialogue. We manually checked dialogues with reported issues and removed those containing major errors, such as when workers misinterpreted the user goal.
\item We deleted dialogues in which the DB query at the end of the dialogue did not match the informable slots in the user goal.
\end{enumerate}
The resulting corpus after this modification contained a total of 4,246 dialogues, Here, the dev and test sets consist of 300 dialogues each, randomly selected from the 4,246 dialogues, and the remaining 3,646 dialogues are used as the train set (refer to Table~\ref{tbl:statistics_of_jmultiwoz} for the statistics of each set).

\subsection{Annotation of Full Dialogue State}
\label{sec:annotation_of_full_dialogue_state}

\begin{table}[t]
\footnotesize
\centering
{\tabcolsep=1.0mm
\begin{tabular}{llp{5cm}} \toprule
Context & User: & I'm looking for a hotel in Sapporo, and it would be helpful if it has free wifi. \\
                  & Wizard: & How about JR INN Sapporo for example? \\
                  & User: & Good, what are the prices like? \\ \midrule
\ml{2}{l}{DB query} & city=Sapporo, wifi=yes \\ \midrule
\ml{2}{l}{Dialogue State} & city=Sapporo, wifi=yes, {\bf name=JR INN Sapporo} \\ \bottomrule
\end{tabular}
}
\caption{Examples of a DB query obtained during dialogue collection and a dialogue state obtained from the additional annotation}
\label{tbl:full_ds_annotation}
\end{table}

\begin{table}[t]
\small
\centering
\begin{tabular}{lr} \toprule
& \textbf{\# slot-value pairs} \\ \midrule
DB search query & 155,274 \\
Newly annotated dialogue state & 58,745 \\
Final dialogue state & 214,019 \\
\bottomrule
\end{tabular}
\caption{Number of slot-value pairs annotated in JMultiWOZ}
\label{tbl:num_slot_values_in_ds}
\end{table}

The dialogue state is the information about the conditions of the entity that the user seeks, known up to each turn and recorded as a set of slot-value pairs. The DB query entered by the wizard at each turn can be used as a part of the dialogue state annotation. However, the DB query does not contain any non-explicitly communicated values. For example, as shown by Table~\ref{tbl:full_ds_annotation}, if the user accepts the entity name proposed by the system (e.g., wizard: ``How about JR INN Sapporo?'', and user: ``OK, what are the prices like?''), the entity name is not searched by the wizard; hence it is not recorded automatically.

To build the complete dialogue state, we recruited additional crowd workers to annotate this non-explicit value. After reading the manual, each worker annotated the values using a dedicated UI. As in the case of dialogue collection, the values were input by selection in order to suppress the perturbation of the slot value notations. To ensure the dialogue quality, each worker annotated ten dialogues for training and received feedback from the authors on errors. After repeated feedback, only those workers who no longer had issues participated in the annotation process.

In the end, six workers shared the annotation tasks for a total of 30,593 wizard turns.\footnote{Of the total 61,186 turns in the dataset (see Table~\ref{tbl:statistics_of_jmultiwoz}), 30,593 were wizard turns, and the dialogue states were annotated for them.} Table~\ref{tbl:num_slot_values_in_ds} shows the statistics of the number of slots in the dialogue states. A total of 58,745 slots (about 37.8\% of the slots recorded in the DB query) were added to the dialogue states. Here, 59 randomly selected dialogues were annotated by the authors, and the match rate between these annotations and those annotated by the workers was 94.1\%, indicating that the annotations were sufficiently high quality.

\begin{table}[t]
\small
\centering
{\tabcolsep=1.3mm
\begin{tabular}{lrrr} \toprule
 & \textbf{MultiWOZ} & \textbf{CrossWOZ} & \textbf{JMultiWOZ} \\ \midrule
Language & English & Chinese & Japanese \\
Domains & 7 & 5 & 6 \\
Slots & 57 & 72 & 79 \\
Dialogues & 8,438 & 5,012 & 3,646 \\
Turns & 115,424 & 84,692 & 52,405 \\
Avg. domains & 1.8 & 3.24 & 2.02 \\
Avg. turns & 13.7 & 16.9 & 14.4 \\
Synchronous & No & Yes & Yes \\ \bottomrule
\end{tabular}
}
\caption{Comparison of multi-domain task-oriented dialogue datasets on the train set; ``Synchronous'' indicates whether one user and one wizard actually conducted the dialogue synchronously during the dialogue collection.}
\label{tbl:comparison_of_datasets}
\end{table}

\subsection{Statistics}
\begin{figure}[t]
\centering
\includegraphics[scale=0.17]{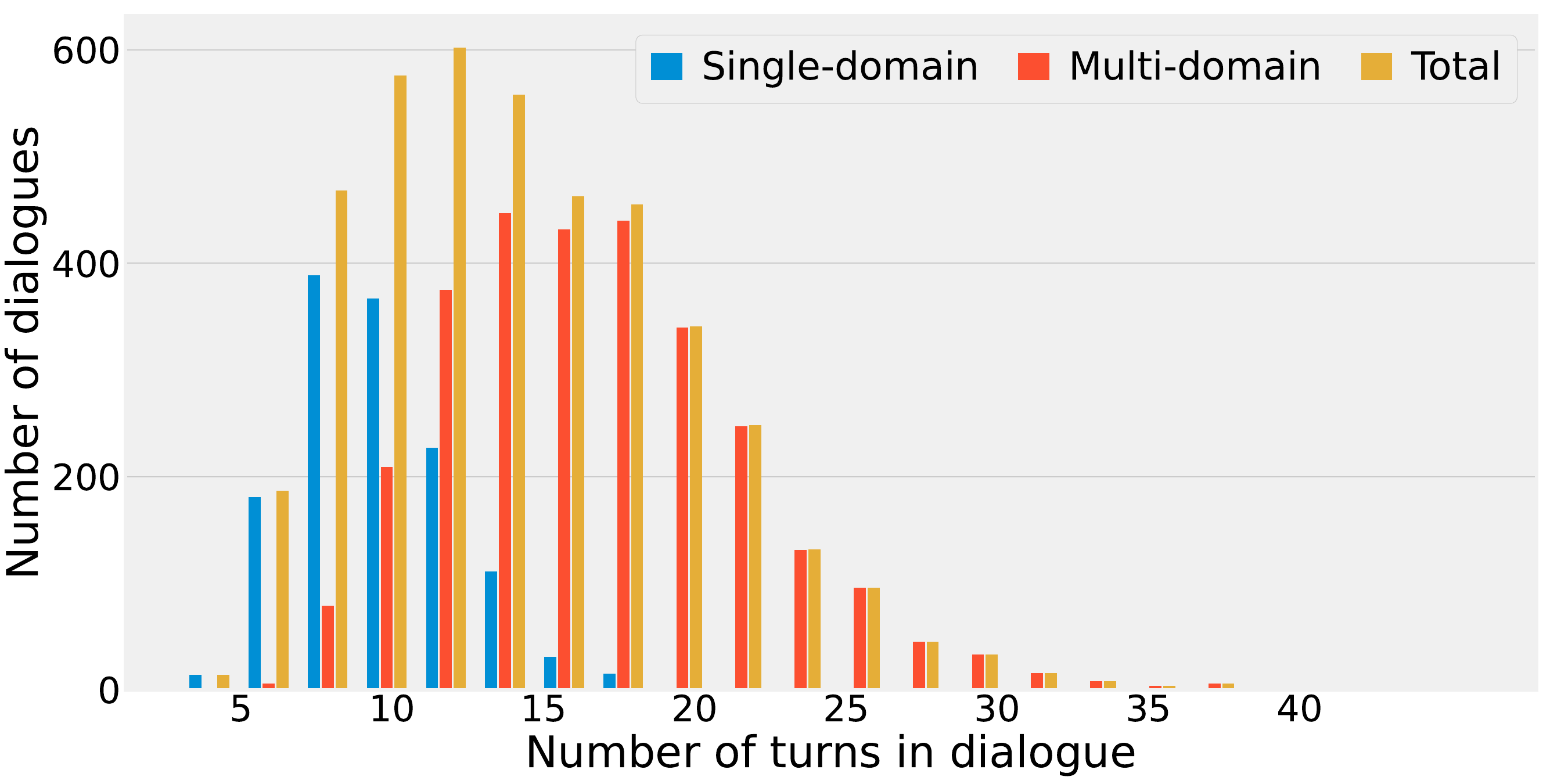}
\caption{The distribution of dialogue lengths, divided into dialogues containing only one domain (single-domain) and dialogues containing two or more domains (multi-domain).}
\label{fig:num_turns_hist}
\end{figure}

Table~\ref{tbl:comparison_of_datasets} shows the statistics of JMultiWOZ compared with the major multi-domain task-oriented dialogue datasets in English and Chinese, MultiWOZ and CrossWOZ, respectively. Given that the number of domains, the number of slots, the average number of domains, and the average number of turns are roughly equivalent, the complexity of dialogues in JMultiWOZ can be considered to be on par with the existing datasets. Figure~\ref{fig:num_turns_hist} illustrates the distribution of the number of dialogue turns when all 4,246 dialogues are divided into dialogues containing only one domain (single-domain) and dialogues containing multiple domains (multi-domain). A wide variety of length and complexity can be seen in both types of dialogues.

\section{Benchmark}
\label{sec:benchmark}

JMultiWOZ provides benchmarks for two common tasks in task-oriented dialogues, dialogue state tracking (DST) and response generation (RG). DST is the task of estimating the dialogue state at each turn, while RG is the task of generating the next system response based on the dialogue context at each turn. To demonstrate that JMultiWOZ can provide benchmarks equivalent to those in existing English dialogue corpora, we evaluate the aforementioned two tasks in JMultiWOZ using the SOTA methods from MultiWOZ2.2~\citep{zang-etal-2020-multiwoz} and the latest LLM-based methods. Note that MultiWOZ2.2 is a version in which various annotation errors in the original MultiWOZ have been corrected.

\begin{figure*}[t]
\begin{minipage}[t]{\linewidth}
\centering
\includegraphics[scale=0.8]{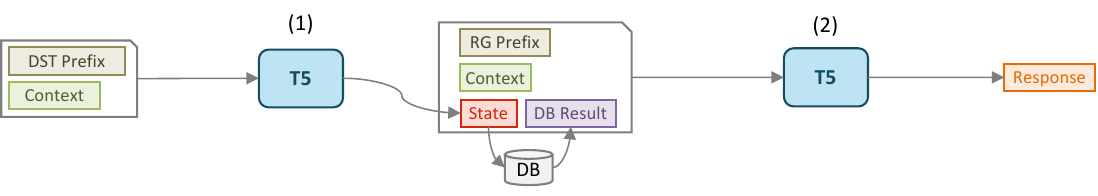}
\subcaption{T5 Pipeline~\citep{bang-etal-2023-task}. (1) First, the dialogue state is predicted from the given dialogue context, and (2) the result is added to the input to generate the final response. Both (1) and (2) are performed on the same model.}
\label{fig:tod_model_t5}
\end{minipage}

\vspace{3mm}

\begin{minipage}[t]{\linewidth}
\centering
\includegraphics[scale=0.8]{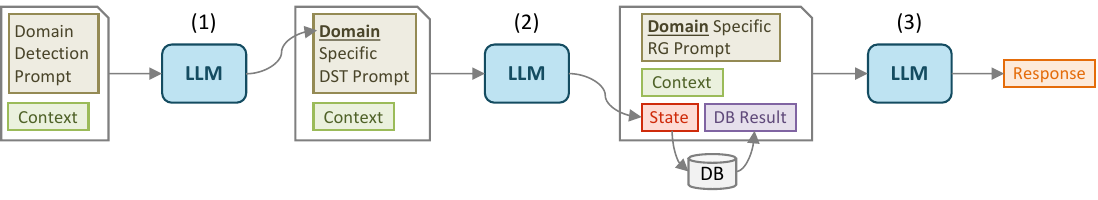}
\subcaption{LLM pipeline~\citep{hudecek-dusek-2023-large} in zero-shot setting. First, (1) the current active domain is estimated from the dialogue context. Next, (2) the dialogue state is tracked, and (3) the response is generated for that domain using a prompt focused on that domain.}
\label{fig:tod_model_chatgpt}
\end{minipage} \\
\caption{Task-oriented dialogue model pipeline for DST and RG}
\label{fig:tod_model}
\end{figure*}

\subsection{Baseline Models}
The most recent task-oriented dialogue models can be generally categorized into two approaches: (1) fine-tuning with supervised learning on medium-sized pretrained language models (Supervised Fine-Tuning; SFT)~\citep{lin-etal-2021-leveraging, su-etal-2022-multi}, and (2) generating responses in a zero-/few-shot setting using an LLM~\citep{hudecek-dusek-2023-large, raposo-etal-2023-prompting}. In this experiment, we validate both methods using the following models. Note that the model-predicted dialogue states were used as input for RG, instead of the ground-truth dialogue states. This is in line with the settings used in previous studies~\citep{bang-etal-2023-task, hudecek-dusek-2023-large}, allowing for comparison with the scores reported in those studies.

\paragraph{Supervised fine-tuning} For SFT, we use the T5~\citep{JMLR:v21:20-074}-based model TOATOD~\citep{bang-etal-2023-task}, which is the SOTA model in MultiWOZ2.2. TOATOD uses T5-base pretrained on a large number of task-oriented dialogue corpora as its backbone model, but such corpora do not exist in Japanese. Therefore, for the evaluation of JMultiWOZ, we simply use T5-base/large\footnote{\url{https://huggingface.co/retrieva-jp/t5-large-long}} pretrained on Japanese data as backbone models.

Figure~\ref{fig:tod_model_t5} depicts the overview of response generation using T5. First, the dialogue state is estimated from the dialogue context (i.e., DST). Then, the database search results, obtained by using the dialogue state as a DB query, are combined with the context and dialogue state to generate the final response (i.e., RG). During fine-tuning, the mapping of ground truth inputs and outputs at each step is learned using maximum likelihood estimation (MLE). Because the same model is used to perform both DST and RG, a prefix indicating the task is added to the input. Specific examples of the input-output sequences used in the training of T5 can be found in Section~\ref{sec:appendix:t5_data_example} in the appendix.

In the training of T5-base and T5-large, the batch size was set to 32, and they were trained for five epochs. The AdamW~\cite{loshchilov2018decoupled} optimizer was used, and the learning rate was initially set to 5e-5 and then linearly decayed according to the number of steps. For evaluation, the checkpoint at the final step was used, and for inference in DST and RG, greedy search was adopted in both cases. 

\begin{table*}[t]
\centering
\begin{tabular}{l|c|ccc|ccc} \toprule
\mr{2}{Model} & \mr{2}{Few-shot} & \ml{3}{c|}{MultiWOZ 2.2} & \ml{3}{c}{JMultiWOZ} \\
 & & JGA & Slot-F1 & BLEU & JGA & Slot-F1 & BLEU \\ \midrule
 T5-base$^\dagger$ & \xmark & 0.64 & 0.94 & 17.04 & 0.59 & 0.95 & 42.31 \\
 T5-large & \xmark & – & – & – & 0.77 & 0.98 & 49.68 \\
 GPT-3.5$^\ddagger$ & \xmark & 0.13 & 0.40 & 4.17 & 0.16 & 0.70 & 5.31 \\
 GPT-4 & \xmark & – & – & – & 0.31 & 0.86 & 8.87 \\ \midrule
 GPT-3.5$^\ddagger$ & \cmark & 0.27 & 0.51 & 6.77 & 0.25 & 0.82 & 12.91 \\
 GPT-4 & \cmark & – & – & – & 0.36 & 0.89 & 15.76 \\ \bottomrule
\end{tabular}
\caption{Automatic evaluation results of the baseline model. Dagger $^\dagger$ indicates that the score of T5-base in MultiWOZ2.2 is cited from the evaluation results of TOATOD reported in \citep{bang-etal-2023-task}. Double-dagger $^\ddagger$ indicates that the score of GPT-3.5 in MultiWOZ2.2 is cited from the reported value in \citep{hudecek-dusek-2023-large}.}
\label{tbl:automatic_evaluation_result}
\end{table*}

\paragraph{LLM zero-/few-shot} We use the zero-/few-shot response generation pipeline using an LLM introduced by \citet{hudecek-dusek-2023-large}. Figure~\ref{fig:tod_model_chatgpt} shows the 3-step flow of zero-shot response generation. First, the LLM estimates the current active domain from the context. Then, focusing on that domain, it estimates the dialogue state. Based on the state, it searches the DB and uses all of the results to generate the final response. In the few-shot setting, two examples retrieved from the train set are mixed into each prompt in the pipeline. These examples are retrieved from the train set based on the similarity between the dialogue context embeddings of the retrieved examples and the current context embeddings. The Japanese sentence-transformers\footnote{\url{https://huggingface.co/cl-nagoya/sup-simcse-ja-large}} were used to create embedding vectors using two consecutive turns of utterances as the dialogue context.

We used APIs of OpenAI GPT-3.5 (\texttt{gpt-3.5-turbo}) and GPT-4 (\texttt{gpt-4}) for the LLM-based zero-/few-shot settings. GPT-3.5 is the highest performing model for DST and RG on MultiWOZ, as reported by \citet{hudecek-dusek-2023-large}. GPT-4 is the latest model provided by OpenAI's API and is known for its high capabilities in Japanese.\footnote{Based on the results of the Japanese LLM leaderboard (\url{https://api.wandb.ai/links/wandb-japan/6ff86bp3} as of October 2023).} The Japanese prompts were created based on the prompts in \citep{hudecek-dusek-2023-large}. See Section~\ref{sec:appendix:prompt_example} in the Appendix for examples of the prompts.

\subsection{Evaluation Metrics}
To evaluate DST, we used the most standard metric, joint goal accuracy (JGA). JGA is a measure of whether the estimated dialogue state perfectly matches the ground truth dialogue state at each turn, resulting in a binary outcome (0/1). Additionally, we also utilized Slot-F1 to evaluate for each turn the match rate between the set of slot values in the estimated dialogue state and the set of slot values in the ground truth state, in terms of F1 score. To evaluate RG, we used the BLEU score, which indicates the similarity between the generated response and the ground truth response.

\subsection{Results}
Table~\ref{tbl:automatic_evaluation_result} shows a comparison between the evaluation results reported in previous studies on MultiWOZ2.2 and those of JMultiWOZ. On JMultiWOZ, the SFT method, namely T5-base/large, generally had the highest performance for both DST and RG, and there seemed to be certain limitations to the performance of LLMs. This trend is consistent with the results of MultiWOZ2.2.

For the DST metrics, Slot-F1 was higher in JMultiWOZ. As described in Section \ref{sec:dialogue_collection}, in the construction of JMultiWOZ, we chose the selection-based input for the DB search query by the wizard to reduce variations in dialogue state reported as issues in MultiWOZ. This enabled the dialogue models to predict exact values without confusion, resulting in higher slot-F1.

For JGA, the primary metric for DST, the difference between MultiWOZ2.2 and JMultiWOZ was minimal. This suggests that the complexity and annotation accuracy of JMultiWOZ are comparable to that of MultiWOZ2.2, demonstrating its capacity to provide benchmarks equivalent to the existing datasets. Notably, there is a difference of about 5\% in JGA for T5-base across both corpora. This gap can be attributed to the fact that in MultiWOZ2.2, besides supervised learning, T5-base was boosted via reinforcement learning with the reward functions where JGA was incorporated.

For the RG metric, BLEU, JMultiWOZ yielded significantly higher scores compared to MultiWOZ2.2. This is likely because the wizard's utterances in JMultiWOZ are more consistent than in MultiWOZ2.2. During the dialogue collection for JMultiWOZ, the dialogues were conducted synchronously, ensuring that the wizards did not switch in the middle of each dialogue. Moreover, we provided ample training for the wizards. Through such quality control measures, we believe that consistent system responses could be attained.

\begin{table*}[t]
\small
\centering
\begin{tabular}{l|cccccc|cccccc} \toprule
\mr{2}{Model} & \ml{6}{c|}{MultiWOZ 2.2} & \ml{6}{c}{JMultiWOZ} \\
& N & Success & Turn & Und. & App. & Sat. & N & Success & Turn & Und. & App. & Sat. \\ \midrule
T5-base$^\dagger$ & 36 & 66.67 & 12.56 & 3.83 & 3.81 & 3.72 & 38 & 65.79 & 10.74 & 3.92 & 3.71 & 3.55 \\
T5-large & – & – & – & – & – & – & 40 & 75.00 & 10.10 & 4.05 & 3.98 & 3.82 \\
GPT-3.5$^\dagger$ & 42 & 57.14 & 11.55 & 3.79 & 3.98 & 4.05 & 41 & \textbf{24.39} & 11.05 & 2.90 & 2.24 & 1.95 \\
GPT-4$^\dagger$ & 42 & 76.19 & 11.88 & 4.26 & 4.36 & 4.00 & 42 & \textbf{57.14} & 9.55 & 3.93 & 3.52 & 3.02 \\ \bottomrule
\end{tabular}
\caption{Human evaluation results. ``N'' indicates the number of participants who interacted with each model. ``Und.'', ``App.'', and ``Sat.'' indicate the worker's subjective evaluation of the system's understanding, appropriateness of the system response, and satisfaction with the dialogue, respectively. Dagger $^\dagger$ indicates that the score in MultiWOZ2.2 is cited from the reported value in \citep{iizuka2023clarifying}.}
\label{tbl:human_evaluation_result}
\vspace{-1mm}
\end{table*}

\section{Human Evaluation}
\label{sec:human_evaluation}
The model's actual task completion capability in real dialogues must be assessed by actual people when evaluating task-oriented dialogue models as well as by automatic evaluation~\citep{iizuka2023clarifying}. In this section, we evaluate the end-to-end dialogue capabilities of the four dialogue models acquired in Section \ref{sec:benchmark}, namely, T5-base, T5-large, GPT-3.5, and GPT-4, by having them engage in conversations with crowd workers. Note that both GPT-3.5 and GPT-4 used a few-shot setting.

\subsection{Settings}
The setting of this evaluation experiment followed that of \citet{iizuka2023clarifying}, where they evaluated the end-to-end dialogue performance of the TOATOD~\citep{bang-etal-2023-task} and \citet{hudecek-dusek-2023-large}'s LLM pipeline models built based on the MultiWOZ2.2 by using crowdsourcing.

Specifically, each worker was first given dialogue instructions and a user goal for each dialogue session. These user goals were randomly sampled from the test set. Then, they engaged in a dialogue with one of the four models. Each dialogue was set to a maximum of 20 turns (one turn consists of one user utterance and one system response), and workers judged whether they achieved the user goal within 20 turns. After the dialogue, the workers evaluated (1) the system's language understanding ability, (2) the appropriateness of the system's responses, and (3) their overall satisfaction with the dialogue, each on a 5-point scale. The workers were restricted to converse with only one system.

\subsection{Results}
Table~\ref{tbl:human_evaluation_result} presents the evaluation results on JMultiWOZ, as well as the results reported by \citet{iizuka2023clarifying} on MultiWOZ2.2. For the models trained with SFT, namely T5-base/large, there was no significant difference in Success, compared to T5-base (TOATOD) on MultiWOZ2.2. These results show that using JMultiWOZ allows for the development and evaluation of Japanese end-to-end dialogue models with performance comparable to that of MultiWOZ2.2.

The performance of the LLM models, i.e., GPT-3.5 and GPT-4, significantly declined compared to that of MultiWOZ2.2. This may be because even the latest LLMs, such as GPT-4, are not capable of handling dynamically changing dialogue contexts in Japanese. Specifically, since the model's predicted dialogue states and system responses are accumulated in the dialogue history, errors propagate gradually, making it difficult to maintain multi-turn conversations. GPT-4's Japanese language ability is not as high as that for English~\citep{OpenAI2023GPT4TR}, and this difference in ability is likely reflected in the performance of task-oriented dialogue systems. This limitation of dialogue skills in LLMs in non-English languages should be addressed with multilingual resources in the future, and we anticipate that JMultiWOZ can contribute to such studies.

\section{Conclusion}
We have presented JMultiWOZ, the first large-scale Japanese multi-domain task-oriented dialogue dataset. This corpus contains 4,246 WOZ dialogues spanning six travel-related domains and provides benchmarks for dialogue state tracking and response generation. Using JMultiWOZ, we evaluated existing SOTA methods with the T5-based models and the latest LLM-based methods, namely GPT-3.5/4, and demonstrated that JMultiWOZ performs comparably to the major benchmark dataset in English, MultiWOZ. Furthermore, we confirmed that the capabilities of GPT-3.5/4 in particular are limited in Japanese.

For future work, we would like to conduct comprehensive evaluation using more diverse models, including other LLMs with high multilingual abilities~\citep{team2023gemini}. We hope that the development of JMultiWOZ will lead to further research on Japanese dialogue systems, including the improvement of LLMs' task-oriented dialogue capabilities in Japanese and the development of a multilingual task-oriented dialogue model.

\section{Limitations}
In this study, to collect native dialogues specific to Japan, we collected online conversations through Japanese crowd workers. However, in prior research, multilingual corpora were built through machine translations of MultiWOZ, and GlobalWOZ~\citep{ding-etal-2022-globalwoz} includes Japanese dialogue data, albeit being entirely machine-translated. Therefore, in the future, it will be necessary to investigate the advantages of JMultiWOZ, which we collected from scratch, compared to GlobalWOZ. Additionally, it may be possible to further improve performance by using GlobalWOZ for pre-training and then fine-tuning with our JMultiWOZ; this will need to be validated in the future.

JMultiWOZ provides benchmarks for two main tasks, DST and RG, through dialogue state annotations, enabling the construction and evaluation of end-to-end task-oriented dialogue models. However, it does not yet have dialogue act (DA) annotations~\citep{budzianowski-etal-2018-multiwoz, eric-etal-2020-multiwoz}. This means that it does not provide a benchmark for another major task in task-oriented dialogue, policy optimization. Therefore, to further enhance its utility, DA annotations should be added in the future.

\section{Ethical Considerations}
Our ethical considerations span across data sources, dialogue collection, human evaluation, and the implications of using pretrained language models (PLMs).

\paragraph{Backend Database} We exclusively employed information sources free from copyright constraints. Our list of entities was primarily derived from websites operated by the government or municipalities. Specific information for each entity was solely extracted from their respective official websites, ensuring authenticity and credibility. We consciously abstained from using tourism sites, or any other source potentially encumbered by copyright issues.

\paragraph{Dialogue Collection and Human Evaluation} Prior to our data collection and evaluation experiments, ethical approval was obtained from the affiliated organization. In the data collection and evaluations, we engaged only crowd workers who explicitly consented to abstain from (1) unsolicited disclosure of personal information during dialogues. Additionally, these workers agreed to (2) relinquish copyright claims over data and artifacts produced during the dialogue collection phase and (3) publication of collected dialogue data. The data to be released will not contain any personally identifiable information of the workers, such as their names.

\paragraph{Dialogue Modeling with PLMs} The pre-training data of the PLMs uses a vast amount of textual data from Internet information. Therefore, our dialogue models based on these PLMs may produce potentially harmful or discriminatory responses.

\section{Acknowledgments}
This work was supported by JST Moonshot R\&D Grant number JPMJMS2011. We used the computational resources of the supercomputer ``Flow'' at the Information Technology Center, Nagoya University.

\section{Bibliographical References}
\label{sec:references}

\bibliographystyle{lrec-coling2024-natbib}
\bibliography{references}

\clearpage
\appendix
\onecolumn

\section{Example of Input-Output Data for T5}
\label{sec:appendix:t5_data_example}
\begin{figure*}[h]
\begin{minipage}{\linewidth}
\centering
\includegraphics[scale=0.85]{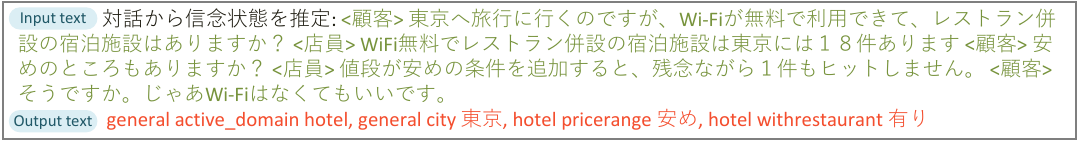}
\subcaption{Original data in Japanese}
\label{fig:t5_dst_data_ja}
\end{minipage}

\vspace{4mm}

\begin{minipage}{\linewidth}
\centering
\includegraphics[scale=0.85]{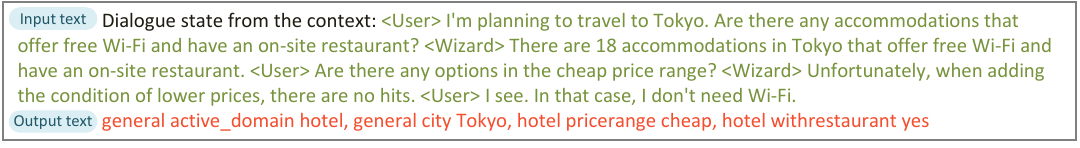}
\subcaption{English translation by the authors}
\label{fig:t5_dst_data_en}
\end{minipage}
\caption{An example of input/output data used for training the T5 on the DST task. At the beginning of the input sequence, a prefix indicating the DST task is attached. Additionally, each element of the input sequence is prefixed with indicators denoting the speaker.}
\label{fig:t5_dst_data}
\end{figure*}

\begin{figure*}[h]
\begin{minipage}{\linewidth}
\centering
\includegraphics[scale=0.85]{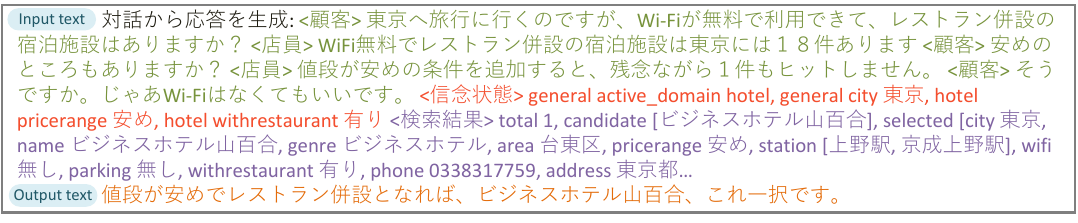}
\subcaption{Original data in Japanese}
\label{fig:t5_rg_data_ja}
\end{minipage}

\vspace{4mm}

\begin{minipage}{\linewidth}
\centering
\includegraphics[scale=0.85]{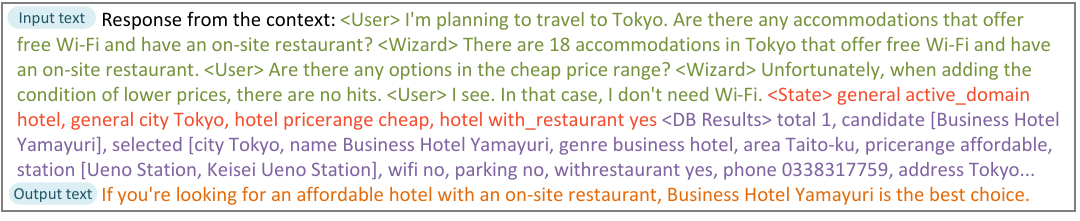}
\subcaption{English translation by the authors}
\label{fig:t5_rg_data_en}
\end{minipage}
\caption{An example of input/output data used for training the T5 on the RG task. At the beginning of the input sequence, a prefix indicating the RG task is attached. Additionally, each element of the input sequence is prefixed with indicators denoting the speaker, belief states, and the results of database searches.}
\label{fig:t5_rg_data}
\end{figure*}

Figures~\ref{fig:t5_dst_data} and \ref{fig:t5_rg_data} show specific examples of the input-output sequences used in the training of T5.

\clearpage

\section{Prompt Examples for LLMs}
\label{sec:appendix:prompt_example}
\begin{figure*}[h]
\begin{minipage}{\linewidth}
\centering
\includegraphics[scale=0.8]{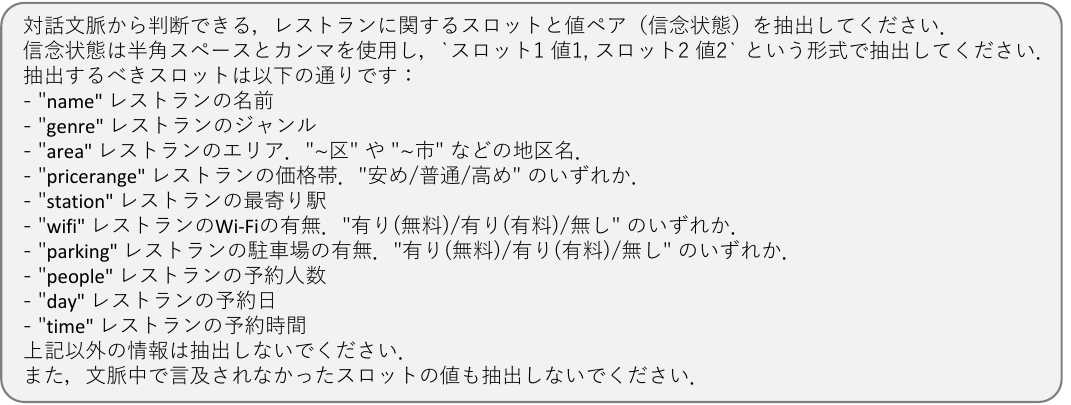}
\subcaption{Original prompt in Japanese}
\label{fig:llm_dst_prompt_ja}
\end{minipage}

\vspace{3mm}

\begin{minipage}{\linewidth}
\centering
\includegraphics[scale=0.8]{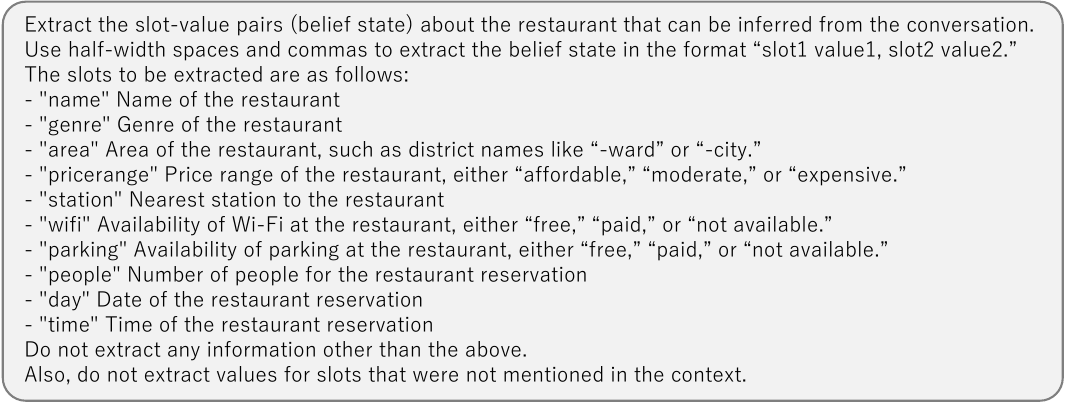}
\subcaption{English translation by the authors}
\label{fig:llm_dst_prompt_en}
\end{minipage}
\caption{Prompt examples used for the DST task in the restaurant domain for the LLMs}
\label{fig:llm_dst_prompt}
\end{figure*}

\begin{figure*}[h]
\begin{minipage}{\linewidth}
\centering
\includegraphics[scale=0.8]{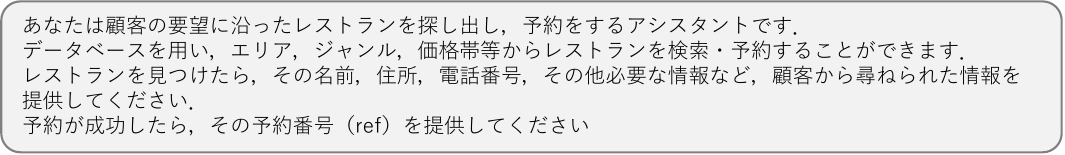}
\subcaption{Original prompt in Japanese}
\label{fig:llm_rg_prompt_ja}
\end{minipage}

\vspace{3mm}

\begin{minipage}{\linewidth}
\centering
\includegraphics[scale=0.8]{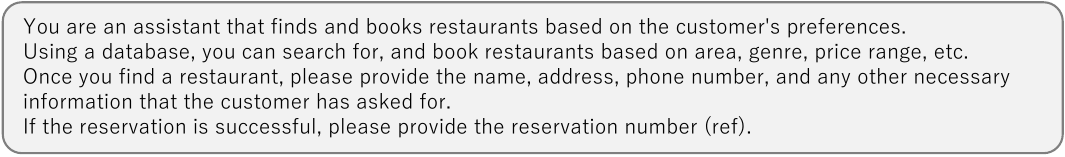}
\subcaption{English translation by the authors}
\label{fig:llm_rg_prompt_en}
\end{minipage}
\caption{Prompt examples used for the RG task in the restaurant domain for the LLMs}
\label{fig:llm_rg_prompt}
\end{figure*}

We created the Japanese prompts for JMultiWOZ based on the prompts for MultiWOZ used by \citet{hudecek-dusek-2023-large}. Figures~\ref{fig:llm_dst_prompt} and \ref{fig:llm_rg_prompt} show examples of prompts used for LLMs in a zero-shot setting. 

\end{document}